\documentclass[conference]{IEEEtran}
\ifCLASSINFOpdf
\else
\fi
\usepackage{amsmath, bm}
\usepackage{amssymb}
\usepackage{xcolor}
\usepackage{graphicx}
\usepackage{hyperref}
\hypersetup{nolinks=true}
\usepackage{cleveref}
\usepackage{makecell}
\usepackage{tabularx}
\usepackage{amsthm}
\usepackage{array}
\usepackage{booktabs}       
\usepackage{multirow}
\usepackage{hyperref}
\usepackage{color}

\usepackage[linesnumbered,ruled,vlined]{algorithm2e}
\usepackage{threeparttable}
\usepackage{subcaption}

\begin{document}

\title{Examining Deep Learning Models with Multiple Data Sources for COVID-19 Forecasting}

\author{\IEEEauthorblockN{Lijing Wang\IEEEauthorrefmark{1}\IEEEauthorrefmark{2},
Aniruddha Adiga\IEEEauthorrefmark{2},
Srinivasan Venkatramanan\IEEEauthorrefmark{2},
Jiangzhuo Chen\IEEEauthorrefmark{2},
Bryan Lewis\IEEEauthorrefmark{2}
and
Madhav Marathe\IEEEauthorrefmark{1}\IEEEauthorrefmark{2}
}
\IEEEauthorblockA{\IEEEauthorrefmark{1}Computer Science, University of Virginia, Charlottesville, VA 22903}
\IEEEauthorblockA{\IEEEauthorrefmark{2}Biocomplexity Institute and Initiative, University of Virginia, Charlottesville, VA 22903\\
Email: \{lw8bn,aa5dw,srini,chenj,brylew,marathe\}@virginia.edu}}

\maketitle
\begin{abstract}
The COVID-19 pandemic represents the most significant public health disaster since the 1918 influenza pandemic. During pandemics such as COVID-19, timely and reliable spatio-temporal forecasting of epidemic dynamics is crucial. Deep learning-based time series models for forecasting have recently gained popularity and have been successfully used for epidemic forecasting. Here we focus on the design and analysis of deep learning-based models for COVID-19 forecasting. We implement multiple recurrent neural network-based deep learning models and combine them using  the stacking ensemble technique. In order to incorporate the effects of multiple factors in COVID-19 spread, we consider multiple sources such as COVID-19 confirmed and death case count data and testing data for better predictions. To overcome the sparsity of training data and to address the dynamic correlation of the disease, we propose clustering-based training for high-resolution forecasting. The methods help us to identify the similar trends of certain groups of regions due to various spatio-temporal effects. We examine the proposed method for forecasting weekly COVID-19 new confirmed cases at county-, state-, and country-level. A comprehensive comparison between different time series models in COVID-19 context is conducted and analyzed. The results show that simple deep learning models can achieve comparable or better performance when compared with more complicated models. We are currently integrating our methods as a part of our weekly forecasts that we provide state and federal authorities. 
\end{abstract}


\section{Introduction}
The COVID-19 pandemic is the worst outbreak we have seen since 1918; it has caused over 22 million confirmed cases globally and over 791,000 deaths in more than 200 countries as of August 26, 2020~\cite{who}. The economic impact is equally staggering, estimates suggest an overall impact of 86.6 trillion U.S. dollars on the global GDP~\cite{duffin2020impact}. One effective way to control epidemics is to forecast the epidemic trajectory -- a good and reliable forecast can
help in planning and response operations.
Two popular methods for forecasting COVID-19 dynamics are statistical time series models and compartmental mass action models at varying spatio-temporal scales ~\cite{anastassopoulou2020data,giordano2020modelling,yamana2020projection,yang2020modified,kai2020universal,chang2020modelling,kerr2020covasim}. There is also recent work on use of DNN and other ML techniques to forecast COVID-19 outbreak~\cite{magri2020first,dandekar2020neural}. 
These methods can make multi-fidelity predictions based on the model resolution. The statistical time series models are popular for their simplicity, while the compartmental models can often capture human decision making and thus provide a path for counterfactual forecasts. Deep learning models are widely used recently for their high forecasting accuracy.
The Centers for Disease Control and Prevention (CDC) COVID-19 forecasting project shows that only one out of 36 teams is using deep learning-based methods for making projections of cumulative and incident deaths and incident hospitalizations due to COVID-19 in the United States~\cite{cdc} as of August 10, 2020. 
The primary challenge for these methods is the lack of training data. 
Other efforts focus on time series-based methodologies to learn patterns in historical epidemic data and other exogenous factors and leverage those patterns for forecasting~\cite{harvey2020time,petropoulos2020forecasting,ribeiro2020short,hu2020artificial,chimmula2020time,arora2020prediction}.
See \cite{volkova2017forecasting,wu2018deep,wang2019defsi,adhikari2019epideep,deng2019graph,wang2020tdefsi} for use of DNNs to forecast epidemic dynamics more broadly.

\medskip
\noindent
\textbf{Our contributions.} \ Our work focuses on exploring \textit{deep learning}-based methods that incorporate \textit{multiple sources} for \textit{weekly 4 weeks ahead} forecasting of COVID-19 new confirmed cases at \textit{multiple geographical resolutions} including country-, state-, and county-level. 
In the context of COVID-19, the problem is more complicated than seasonal influenza forecasting for the following reasons: 
($i$) very sparse training data for each region;
($ii$) noisy surveillance data due to heterogeneity in epidemiological context e.g. disease spreading timeline and testing prevalence in different regions, ($iii$) system is constantly in churn -- individual behavioral adaptation, policies and disease dynamics are constantly co-evolving.
Given these challenges, we examine different types of time series models and propose an ensemble framework that combines simple deep learning models using multiple sources such as COVID-19 cases data and testing data. The multi-source data allows us to capture the above mentioned factors more effectively.
To overcome the data sparsity problem we propose clustering-based training methods to augment training data for each region. 
We group spatial regions based on trend similarity and infer a model per cluster. Among other things this avoids overfitting due to sparse training data. As an additional benefit it aids in explicitly uncovering the spatial correlation across regions by training models with similar time series. Our main contributions are summarized below:

\begin{itemize}
\item First, we systematically examine time series-based deep learning models for COVID-19 forecasting and propose clustering-based training methods to augment sparse and noisy training data for high resolution regions which can avoid overfitting and explicitly uncover the similar spreading trends of certain groups of regions.

\item Second, we implement a stacking ensemble framework to combine multiple deep learning models and multiple sources for better performance. Stacking is a natural way to combine multiple methods and data sources.

\item Third, we analyze the performance of our method and other published results in their ability to forecast weekly new confirmed cases at country, state, and county level. The results show that our ensemble model outperforms any individual models as well as several classic machine learning and state-of-the-art deep learning models; 

\item Finally, we conduct a comprehensive comparison among mechanistic models, statistical models and deep learning models. The analysis shows that for COVID-19 forecasting deep learning-based models can capture the dynamics and have better generalization capability as opposed to the mechanistic and statistical baselines. Simple deep learning models such as simple recurrent neural networks can achieve better performance than complex deep learning models like graph neural networks for high resolution forecasting. 

\end{itemize}

\section{Related work}
COVID-19 is a very active area of research and thus it is impossible to cover all the recent manuscripts. We thus only cover
important papers here. 

\subsection{COVID-19 forecasting by mechanistic methods}
Mechanistic methods have been a mainstay for COVID-19 forecasting due to their capability of represent the underlying disease transmission dynamics as well as incorporating diverse interventions. They enable counterfactual forecasting which is important for future government interventions to control the spread. Forecasting performance depends on the assumed underlying disease model. 
Yang et al.~\cite{yang2020modified} use a modified susceptible(S)-exposed(E)-infected(I)-recovered(R) (SEIR) model for predicting the COVID-19 epidemic peaks and sizes in China.
Anastassopoulou et al.~\cite{anastassopoulou2020data} provide estimations of the basic reproduction number and the per day infection mortality and recovery rates using an susceptible(S)-infected(I)-dead(D)-recovered(R) (SIDR) model.
Giordano et al.~\cite{giordano2020modelling} propose a new susceptible(S)-infected(I)-diagnosed(D)-ailing(A)-recognized(R)-threatened(T)-healed(H)-extinct(E) (SIDARTHE) model to help plan an effective control strategy. 
Yamana et al.~\cite{yamana2020projection} use a metapopulation SEIR model for US county resolution forecasting.
Chang et al.~\cite{chang2020modelling} develop an agent-based model for a fine-grained computational simulation of the ongoing COVID-19 pandemic in Australia.
Kai et al.~\cite{kai2020universal} present a stochastic dynamic network-based compartmental SEIR model and an individual agent-based model to investigate the impact of universal face mask wearing upon the spread of COVID-19.

\subsection{COVID-19 forecasting by time series models}
Time series models, such as statistical models and deep learning models, are popular for their simplicity and forecasting accuracy in the epidemic domain. One big challenge is the lack of sufficient training data in the context of COVID-19 dynamics. Another challenge is that the surveillance data is extremely noisy (hard to model noise) due to rapidly evolving epidemics.  However, additional data becomes available and the surveillance systems mature these models become more promising.
Harvey et al.~\cite{harvey2020time} propose a new class of time series models based on generalized logistic growth curves that reflect COVID-19 trajectories.
Petropoulos et al.~\cite{petropoulos2020forecasting} produce forecasts using models from the exponential smoothing family.
Ribeiro et al.~\cite{ribeiro2020short} evaluate multiple regression models and stacking-ensemble learning for COVID-19 cumulative confirmed cases forecasting with one, three, and six days ahead in ten Brazilian states.
Hu et al.~\cite{hu2020artificial} propose a modified auto-encoder model for real-time forecasting of the size, lengths and ending time in China.
Chimmula et al.~\cite{chimmula2020time} use LSTM networks to predict COVID-19 transmission.
Arora et al.~\cite{arora2020prediction} use LSTM-based models for positive reported cases for 32 states and union territories of India.
Magri et al.~\cite{magri2020first} propose a data-driven model trained with both data and first principles.
Dandekar et al.~\cite{dandekar2020neural} use neural network aided quarantine control models to estimate the global COVID-19 spread.

\subsection{Deep learning-based epidemic forecasting}
Recurrent neural networks (RNN) has been demonstrated to be able to capture dynamic temporal behavior of a time sequence. Thus it has become a popular method in recent years for seasonal influenza-like-illness (ILI) forecasting.
Volkova et al.~\cite{volkova2017forecasting} build an LSTM model for short-term ILI forecasting using CDC ILI and Twitter data. 
Venna et al.~\cite{venna2019novel} propose an LSTM-based method that integrates the impacts of climatic factors and geographical proximity.
Wu et al.~\cite{wu2018deep} construct CNNRNN-Res combining RNN and convolutional neural networks to fuse information from different sources.
Wang et al.~\cite{wang2019defsi,wang2020tdefsi} propose TDEFSI combining deep learning models with casual SEIR models to enable high-resolution ILI forecasting with no or less high-resolution training data.
Adhikari et al.~\cite{adhikari2019epideep} propose EpiDeep for seasonal ILI forecasting by learning meaningful representations of incidence curves in a continuous feature space.
Deng et al.~\cite{deng2019graph} design cola-GNN which is a cross-location attention-based graph neural network for forecasting ILI.
Regarding COVID-19 forecasting, Amol et al.~\cite{kapoor2020examining} examined a novel forecasting approach for COVID-19 daily case prediction that uses graph neural networks and mobility data. Gao et al.~\cite{gao2020stan} proposed STAN that uses a spatio-temporal attention network.  
Aamchandani et al.~\cite{ramchandani2020deepcovidnet} presented DeepCOVIDNet to compute equidimensional representations of multivariate time series. 
These works examine their models on daily forecasting for US state or county levels.

Our work focuses on time series deep learning models for COVID-19 forecasting that yield weekly forecast at multiple resolution scales and  
provide 4 weeks ahead forecasts (equal to 28 days ahead in the context of daily forecasting). We use an ensemble model to combine multiple simple deep learning models. We show that compared to state-of-the-art time series models, simple recurrent neural network-based models can achieve better performance. More importantly, we show that the ensemble method is an effective way to mitigate model overfitting caused by the super small and noisy training data.
\begin{figure*}[!t]
\centering
\includegraphics[width=0.8\textwidth]{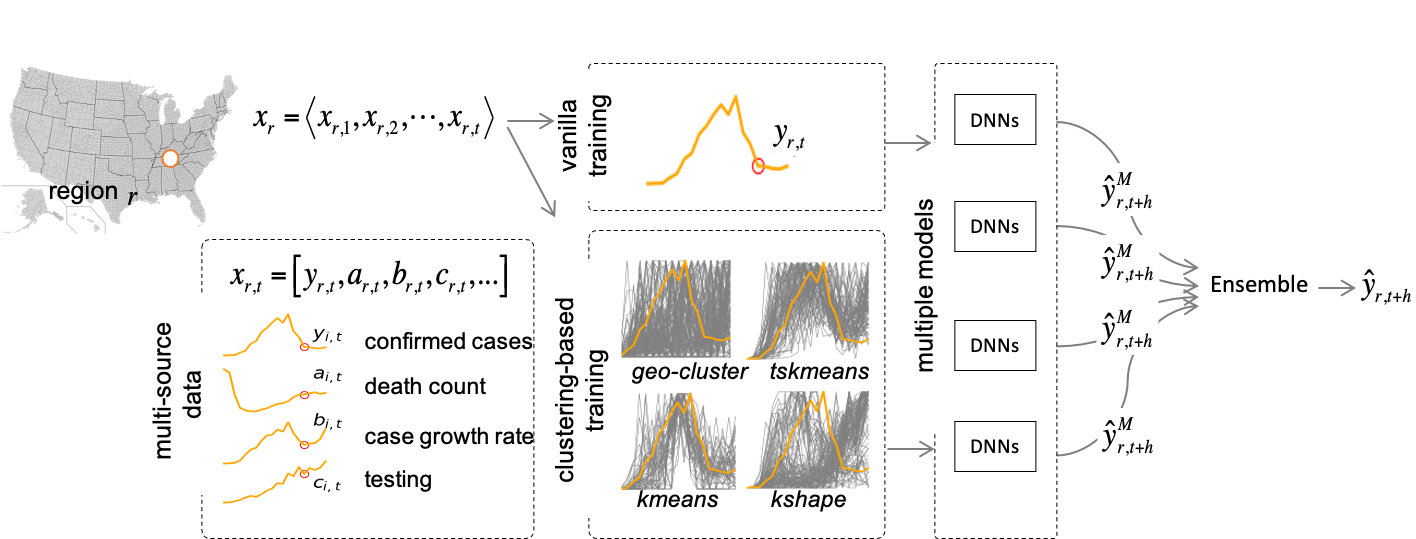}
  \caption{Framework of deep learning based multi-source ensemble.}
  \label{fig:framework}
\end{figure*}

\section{Method}
\subsection{Problem Formulation}
We formulate the COVID-19 new confirmed cases forecasting problem as a regression task with time series of multiple sources as the input. We have $N$ regions in total. Each region is associated with a time series of multi-source input in a time window $T$. For a region $r$, at time step $t$, the multi-source input is denoted as $\mathbf{x}_{r,t} \in \mathbb{R}^S$ where $S$ is the feature number. We denote the training data as $\mathbf{X}_{r,t}=[\mathbf{x}_{r,t-T+1},...,\mathbf{x}_{r,t}] \in \mathbb{R}^{T \times S}$. The objective is to predict COVID-19 new confirmed cases at a future time point $t+h$ where $h$ refers to the horizon of the prediction. 
We are interested in a predictor $f$ that predicts new confirmed case count at time $t+h$, denoted as $\mathbf{z}_{r,t+h}$, by taking $\mathbf{X}_{r,t}$ as the input where $t$ is the most recent time of data availability.
\begin{equation}
\label{equ:predictor}
\hat{\mathbf{z}}_{r,t+h} = f(\mathbf{X}_{r,t}, \theta)
\end{equation}
where $\theta$ denotes parameters of the predictor and $\hat{\mathbf{z}}_{r,t+h}$ denotes the prediction of $\mathbf{z}_{r,t+h}$.

\subsection{Recurrent Neural Networks (RNNs)}
For brevity, we assume a region is given, thus we omit subscript $r$ in this subsection. 
An RNN model consists of k-stacked RNN layers.
Each RNN layer consists of $T$ cells, denoted as $\langle cell_{t-T+1},\cdots,cell_{t} \rangle$. The input is $\mathbf{X}_t$, the output from the last layer $k$ is denoted as $\mathbf{h}^{(k)}$. 
Let $H^{(i)}, 1 \leq i \leq k$ be the dimension of the hidden state in $layer_i$. For the first layer $layer_1$, $cell_t$ will work as:
\begin{equation}
\label{equ:rnncell}
\begin{aligned}
\mathbf{h}_{t}^{(1)} = \tanh(\mathbf{W}_i^{(1)} \cdot \mathbf{x}_{t} + \mathbf{U}_i^{(1)} \cdot \mathbf{h}_{t-1}^{(1)} + \mathbf{b}_i^{(1)}) \in \mathbb{R}^{H^{(1)}}\\
\end{aligned}
\end{equation}
where $\tanh$ is activation function; $\mathbf{W} \in \mathbb{R}^{H^{(1)}\times S}, \mathbf{U} \in \mathbb{R}^{H^{(1)} \times H^{(1)}}$, and $\mathbf{b} \in \mathbb{R}^{H^{(1)}}$ are learned weights and bias; $\mathbf{h}_{t}^{(1)}$ is the output of $cell_t$ and $\mathbf{h}_{t-1}^{(1)}$ is from $cell_{t-1}$. The cell computation is similar in the $layer_i$, but with $\mathbf{x}_{t}$ being replaced by $\mathbf{h}_{t}^{(i-1)} \in \mathbb{R}^{H^{(i-1)}}$, and $\mathbf{W} \in \mathbb{R}^{H^{(i)} \times H^{(i-1)}}$. 
The first RNN layer takes $\mathbf{x}_{t-T+1},\cdots,\mathbf{x}_{T}$ as the input, the second layer takes $\mathbf{h}_{t-T+1}^{(1)},...,\mathbf{h}_{t}^{(1)}$ as the input, and the rest of the layers behave in the same manner.
The RNN module can be replaced by Gated Recurrent Unit (GRU)~\cite{cho2014learning} or Long Short-term Memory (LSTM)~\cite{hochreiter1997long} which avoid short-term memory and gradient vanishing problems of vanilla RNNs.

The output of the k-stacked RNN layers is fed into a fully connected layer:
\begin{equation}
\label{equ:output-1}
\hat{\mathbf{z}}_t = \psi(\mathbf{w} \cdot \mathbf{h}_{t}^{(k)} + \mathbf{b}) \in \mathbb{R}^{H}
\end{equation}
where $H$ is the output dimension, $\mathbf{w} \in \mathbb{R}^{H \times H^{(k)}}$, $\mathbf{b} \in \mathbb{R}^{H}$, and $\psi$ is a linear function.

\subsection{Multi-source Attention RNNs}
The Multi-source attention RNN model consists of $m$ k-stacked RNN models, each of which encodes a time series of one feature. Assume the output of branch $r$ is $\mathbf{h}^r \in \mathbb{R}^{H_r}$ in which we omit subscript $t$ for brevity.
An attention layer is used to measure the impact of multi-source on new confirmed cases.  We assume the time series of new confirmed cases is encoded in branch $r$, and we define attention coefficient $a_j$ as the effect of feature $j$ on target feature:
\begin{equation}
\label{equ:attention}
a_j = \psi (\mathbf{w}_r^T \cdot \mathbf{h}^r + \mathbf{w}_j^T \cdot \mathbf{h}^j + b_j) \in \mathbb{R}
\end{equation}
where $\mathbf{w}_r \in \mathbb{R}^{H_r}$, $\psi$ is RELU function. Then the output of attention layer is: 
\begin{equation}
\label{equ:attention-out}
\mathbf{h}^a = \psi (\mathbf{w}_a\sum_{j=1}^{m}a_j \cdot \mathbf{h}^j + \mathbf{b}_a) \in \mathbb{R}^{H_a}
\end{equation}
where $\mathbf{w}_a \in \mathbb{R}^{H_a \times H_r}$, $\mathbf{b}_a \in \mathbb{R}^{H_a}$, $\psi$ is the tanh function.
The output layer is a dense layer that outputs $\hat{\mathbf{z}}_t$:
\begin{equation}
\label{equ:output-2}
\hat{\mathbf{z}}_t = \psi (\mathbf{w}_o \cdot \mathbf{h}^a + \mathbf{b}_o) \in \mathbb{R}^{H}
\end{equation}
where $\mathbf{w}_o \in \mathbb{R}^{H \times H_a}$, $\mathbf{b} \in \mathbb{R}^{H}$, $\psi$ is the linear function.
In our paper, all the features have the same length of time series. However, the multi-source attention RNN model enables training with the input that has a different length of time series of the features, which is superior in heterogeneous availability of multiple factors.

\subsection{Clustering-based Training}
Deep learning models usually require a large amount of training data which is not the case in the context of COVID-19. Particularly, for regions where the pandemic starts late, there are only a few valid data points for weekly forecasting. Thus training a single model for each such region, which we call \textit{vanilla} training, is highly susceptible to overfitting. One modeling strategy is to train a model for a group of selected regions which to some extent overcomes the data sparsity problem.
It is more likely that groups of regions exhibit strong correlations due to various spatio-temporal effects and geographical or demographic similarity. We explore a clustering-based approach that simultaneously learns COVID-19 dynamics from multiple regions within the cluster and infers a model per cluster. Various types of similarity metrics can be used to uncover 
the trend similarity allowing for an explainable time series forecasting framework.

Generalizing the earlier problem formulation, we denote the historical available time series for a region $r$ as $\mathbf{X}_{r}=[\mathbf{x}_{r,1},...,\mathbf{x}_{r,T_r}] \in \mathbb{R}^{T_r \times S}$ where $T_r$ is the time span of the available surveillance data. $T_{r}$ is increasing as new data becomes available and it varies across different regions. 
The set of time series for $N$ regions is denoted as $\mathcal{X}=\{\mathbf{X}_{r}|r=1,\cdots,N\}$.     
The clustering process aims to partition the $\mathcal{X}$ into $k (\leq N)$ sets $\mathcal{C} = \{C_1,\dots,C_k\}$.

In our work, the trend is represented as the time series of new confirmed cases and we cluster the time series in two ways -- geography-based clustering (\textit{geo-clustering}) and algorithm-based clustering (\textit{alg-clustering}). 
\textit{Geo-clustering}: Clustering is based on their geographical proximity, e.g. partition counties $\mathcal{X}$ based on their state codes for the US. We propose this method due to differences across regions with respect to their size, population density, epidemiological context, and differences in how policies are being implemented. Thus we assume those who belong to the same jurisdictions would have strong relationship in COVID-19 time series. 
\textit{Alg-clustering}: Clustering using (\textit{i}) k-means \cite{hartigan1979algorithm} which partitions $N$ observations into $k$ clusters in which each observation belongs to the cluster with the nearest mean; (\textit{ii}) time series k-means (tskmeans) \cite{huang2016time} that clusters time series data using the smooth subspace information; (\textit{iii}) kshape \cite{paparrizos2015k}  uses a normalized version of the cross-correlation measure in order to consider the shapes of time series while comparing them. Note that kmeans requires the time series to be clustered must have the same length, while geo-clustering, tskmeans and kshape allow for clustering on different lengths of time series. 
Alg-clustering discovers implicit correlation of epidemic trends which does not assume any geographical knowledge. 
We denote the set of above methods as $\mathcal{A}=\{A_{vani},A_{geo},A_{km},A_{ts},A_{ks}\}$. 

\subsection{Ensemble}
Ensemble learning is primarily used to improve the model performance. Ren et. al.~\cite{ren2016ensemble} present a comprehensive review. In this paper, we implement \textit{stacking} ensemble. It is to train a separate dense neural network using the predictions of individual models as the inputs. We use leave-one-out cross validation to train and predict for each region. For each target value $\mathbf{z}_t$, we train the ensemble model using the training samples from the same region but other time points. 

\subsection{Probabilistic Forecasting}
In the epidemic forecasting domain, probabilistic forecasting is important for capturing the uncertainty of the disease dynamics and to better support public health decision making.
We implement MCDropout~\cite{gal2016dropout} for each individual predictors to demonstrate estimation of prediction uncertainty. However, the ensemble predictions are point estimation by the definition of stacking. 

\subsection{Proposed Framework}
Fig.~\ref{fig:framework} shows the framework of the proposed method. It works as follows: (1) we choose a geographical scale and resolution, e.g. counties in the US; (2) we collect and process multi-source training data; (3) we cluster regions into certain groups based on their similarities between time series of new confirmed cases; (4) we train multiple predictors per cluster and ensemble individual predictors to make final predictions. 
\subsubsection{Multiple data sources}
In order to model the co-evolution of multiple factors in COVID-19 spread, we incorporate the following data sources in our models to make future forecasts.  
\textit{COVID-19 Surveillance Data}~\cite{uva2020uva} and \textit{Case Count Growth Rate (CGR)} quantify case count and case count changes of COVID-19 time series.
\textit{COVID-19 Testing Data}~\cite{jhu2020covid}, \textit{Testing Rate (TR)} and \textit{Testing Positive Rate (TPR)} quantify the COVID-19 testing coverage in each region. 
We denote the set of multiple sources as $\mathcal{D}$ where $\mathcal{D}$ can be expanded by combining any new data sources. 
We generate $\mathcal{X}$ by preprocessing $\mathcal{D}$.
Details of data description and generation are shown in section~\ref{subsec:data}.

\subsubsection{Multiple RNN-based models}
By combining different data sources (single feature, $m$ features, attention $m$ features), RNN modules (RNN, GRU, LSTM), and training methods (vanilla, geo, kmeans, tskmeans, kshape), we implement multiple individual models. 
For country, US state and US county levels, models include:
\textit{RNN}, \textit{GRU}, \textit{LSTM} use vanilla training with single feature;
\textit{RNN-m}, \textit{GRU-m}, \textit{LSTM-m} use vanilla training with $m$ features;
\textit{RNN-att}, \textit{GRU-att}, \textit{LSTM-att} are attention-based models using vanilla training with $m$ features.
For US county level, to investigate the effect of clustering training, we implement additional models using RNN module and single feature: \textit{RNN-geo}, \textit{RNN-kmeans}, \textit{RNN-tskmeans} and \textit{RNN-kshape}. We analyze the effects by varying clustering methods while fixing other factors. Thus other combinations of modules, features and training methods are omitted in this work.  
We denote the set of individual models as $\mathcal{M}$.
Note that $\mathcal{M}$ is not limited to the models we implemented in this paper. It can be expanded by adding or improving upon any of the individual components. 

\subsubsection{Training and forecasting}
Algorithm~\ref{alg:framework} presents how the proposed framework works. We first preprocess the collected data sources $\mathcal{D}$ to generate $\mathcal{X}$ based on the data availability for different resolutions. Each feature is in the form of time series of weekly data points at a given geographical resolution. We design various models $\mathcal{M}$ for different resolutions based on $\mathcal{D}$. Next, each model in $\mathcal{M}$ is trained using its corresponding cluster of training data. For region $r$, given an input $\mathbf{X}_{r,t}$, a model $M$ will output $\mathbf{\hat{z}}_{r,t+h}^{M}$. Then the outputs of individual models in $\mathcal{M}$ will be combined using stacking ensemble which will output the final prediction $\mathbf{\hat{z}}_{r,t+h}$ for region $r$ at time $t+h$.

\newlength\mylen
\newcommand\myinput[1]{%
  \settowidth\mylen{\KwIn{}}%
  \setlength\hangindent{\mylen}%
  \hspace*{\mylen}#1\\}

\SetKwInput{KwInput}{Input}
\SetKwInput{KwOutput}{Output}
\begin{algorithm}
    \KwInput{$\mathbf{X}_{r,t}$: the input time series for region $r$; $\mathbf{X}_{r}$: historical time series for region $r$; $\mathcal{X}$: the set of time series for $N$ regions; $\mathcal{A}$: the set of clustering methods; $\mathcal{M}$: the set of model types; }
    \KwOutput{$\mathbf{\hat{z}}_{r,t+h}$: new confirmed case forecasting at time $t+h$}
    \KwData{$\mathcal{D}$: the set of data sources}
    Preprocess $\mathcal{D}$ to generate $\mathcal{X}$ and $\mathbf{X}_{r,t}$\\
    $O_r \leftarrow \emptyset$ \tcp{The set of individual model predictions}
    \For{$A$ in $\mathcal{A}$}{ 
        $C_r \leftarrow \{\mathbf{X}_r\}$  
        
        \tcc{Start clustering}
        $C_A \leftarrow A.fit(\mathcal{X})$ \tcp{$C_A$ is the clustering results using method $A$}
        \For{$i$ in $1,\dots,N$}{
            \If{$\mathbf{X}_r$ and $\mathbf{X}_i$ belong to the same cluster in $C_A$}{
            $C_r := C_r \cup \mathbf{X}_i$}
        }
        \tcc{Start training, forecasting}
        \For{$M$ in $\mathcal{M}$}{
            \If{$M$ is $A$ related}{
            train $M$ using $C_r$\\
            $\mathbf{\hat{z}}_{r,t+h}^{M} := M(\mathbf{X}_{r,t},\theta)$\\
            $O_r := O_r \cup \mathbf{\hat{z}}_{r,t+h}^{M}$
            }
        }
    }
    $\mathbf{\hat{z}}_{r,t+h} := \mathcal{F}(O_r,w)$ \tcp{$\mathcal{F}$ is ensemble algorithm}
\caption{Pseudocode of the proposed framework}
\label{alg:framework}
\end{algorithm}

\subsubsection{Multi-step forecasting}
For single feature, we use a recursive forecasting approach to make multi-step forecasting. That is appending the most recent prediction to the input for the next step forecasting. For multiple features that include exogenous time series as the input, we train a separate model for each step ahead forecasting.
\section{Experiment Setup}

\subsection{Data}\label{subsec:data}
\begin{itemize}
\item \textbf{COVID-19 surveillance data} is obtained via the UVA COVID-19 surveillance dashboard \cite{uva2020uva}. It contains daily confirmed cases (\textbf{CF}) and death count (\textbf{DT}) at the resolution of county/state in the US and national-level data for other countries. Daily case counts and death counts are further aggregated to weekly counts. 

\item \textbf{Case count growth rate (CGR)}: Denoting the new confirmed/death case count at week $t$ as $n_t$, the CGR of week $t+1$ is computed as $log(n_{t+1}+1)-log(n_t+1)$, where we add 1 to smooth zero counts. We compute confirmed CGR (\textbf{CCGR}) and death CGR (\textbf{DCGR}). 

\item \textbf{COVID-19 testing data} via JHU COVID-19 tracking project \cite{jhu2020covid}. It includes multiple data like positive and negative testing count for state and country level of the US. We compute testing per 100K (\textbf{TR}) and testing positive rate (\textbf{TPR}) i.e. positive/(positive+negative).

\end{itemize}

All data sources are weekly and ends on Saturday. It starts from Week ending March 7th and ends at Week ending August 22nd (25 weeks) at \textbf{Global}, \textbf{US-State} and \textbf{US-County} resolutions. The global dataset includes Austria, Brazil, India, Italy, Nigeria, Singapore, the United Kingdom, and the United States. The summary of each dataset is shown in Table~\ref{tab:data}. We chose 2020/03/07 as the start week since commercial laboratories began testing for SARS-CoV-2 in the US on March 1st, 2020. Thus the COVID-19 surveillance data before that date is substantially noisy. The forecasting week starts from 2020/05/23 and we make 4 weeks ahead forecasting at each week until 2020/08/22. For example, if we use time series of data from 2020/03/07 to 2020/05/16 to train models, then the forecasting weeks are 2020/05/23, 2020/05/30, 2020/06/06, and 2020/06/13. Then we move one week ahead to repeat the training and forecasting. 

\begin{table}[t]
\centering
\caption{Dataset Summary. }\label{tab:data}
\begin{tabular}{lccc}
\toprule
\textbf{Data set} &\textbf{\# regions}  &\textbf{\# weeks} &\textbf{\# features} \\
\midrule
Global &8 &25 &6 \\
US-State &50 &25 &8 \\
US-County &2952 &25 &7 \\
\bottomrule
\end{tabular}
\label{space}
\end{table}

\subsection{Metrics}
The metrics used to evaluate the forecasting performance are: \textit{root mean squared error (RMSE)}, \textit{mean absolute percentage error (MAPE)}, \textit{Pearson correlation (PCORR)}.

\begin{itemize}
\item \textbf{Root mean squared error ($\mathbf{RMSE}$)}: 
\begin{equation}
\label{equ:rmse}
\text{RMSE} = \sqrt{\frac{1}{n}\sum_{i=1}^{n}(z_i - \hat{z_i})^2}
\end{equation}

\item \textbf{Mean absolute percentage error ($\mathbf{MAPE}$)}:
\begin{equation}
\label{equ:mape}
\text{MAPE} = (\frac{1}{n}\sum_{i=1}^{n}|\frac{z_i - \hat{z_i}}{z_i+1}|)*100
\end{equation}

\item \textbf{Pearson correlation ($\mathbf{PCORR}$)}: 
\begin{equation}
\label{equ:pcorr}
\text{PCORR}={\frac {\sum _{i=1}^{n}(\hat{z}_{i}-{\bar {\hat{z}}})(z_{i}-{\bar {z}})}{{\sqrt {\sum _{i=1}^{n}(\hat{z}_{i}-{\bar {\hat{z}}})^{2}}}{\sqrt {\sum _{i=1}^{n}(z_{i}-{\bar {z}})^{2}}}}}
\end{equation}
\end{itemize}

   \begin{figure*}[!t]
  \centering 
   \begin{subfigure}[b]{.3\textwidth}
    \centering
    \includegraphics[width=1\textwidth]{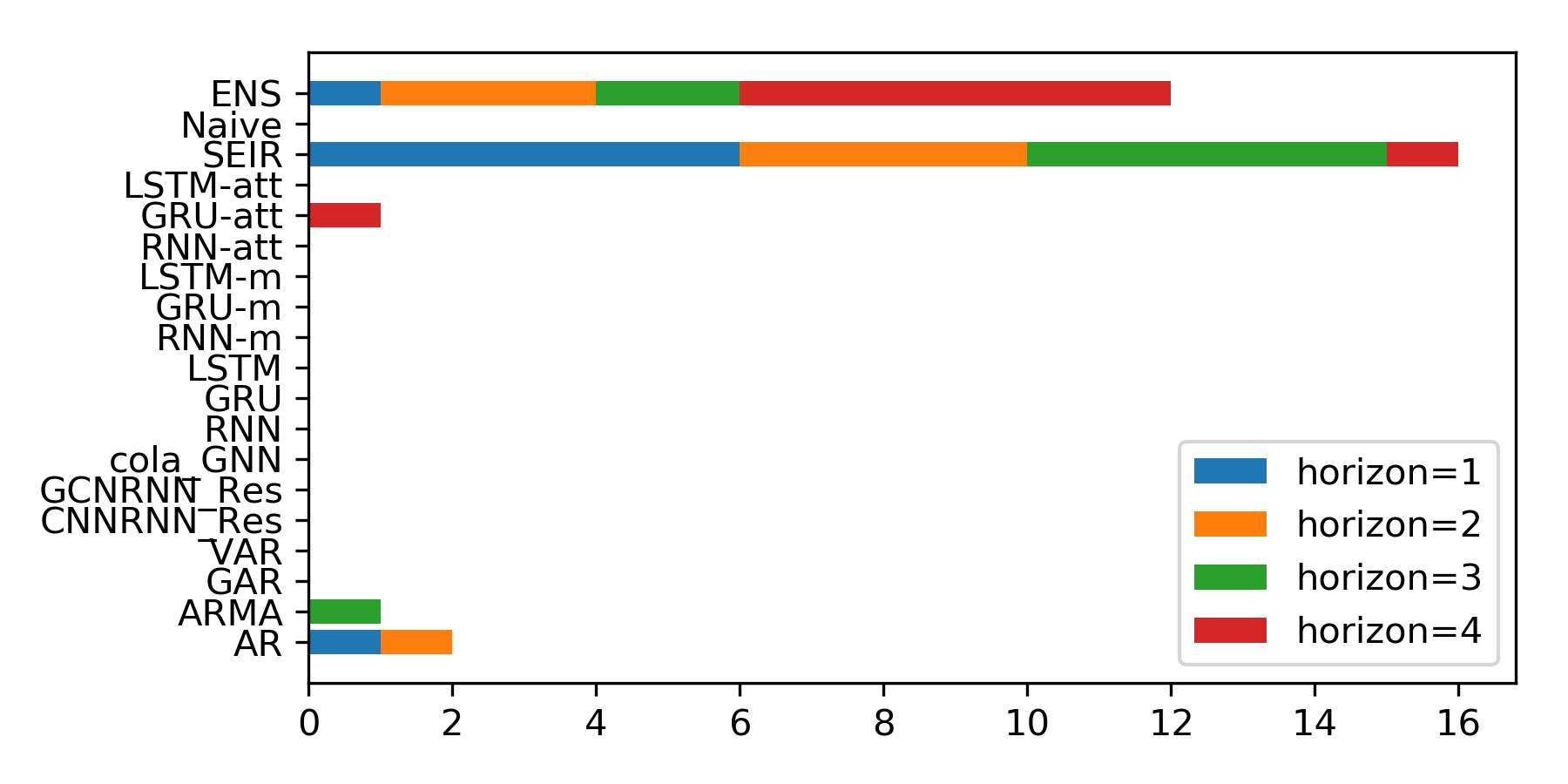}
    \subcaption{Global}
    \label{fig:bar-global}
  \end{subfigure} 
  \quad
  \begin{subfigure}[b]{.3\textwidth}
    \centering
    \includegraphics[width=.97\textwidth]{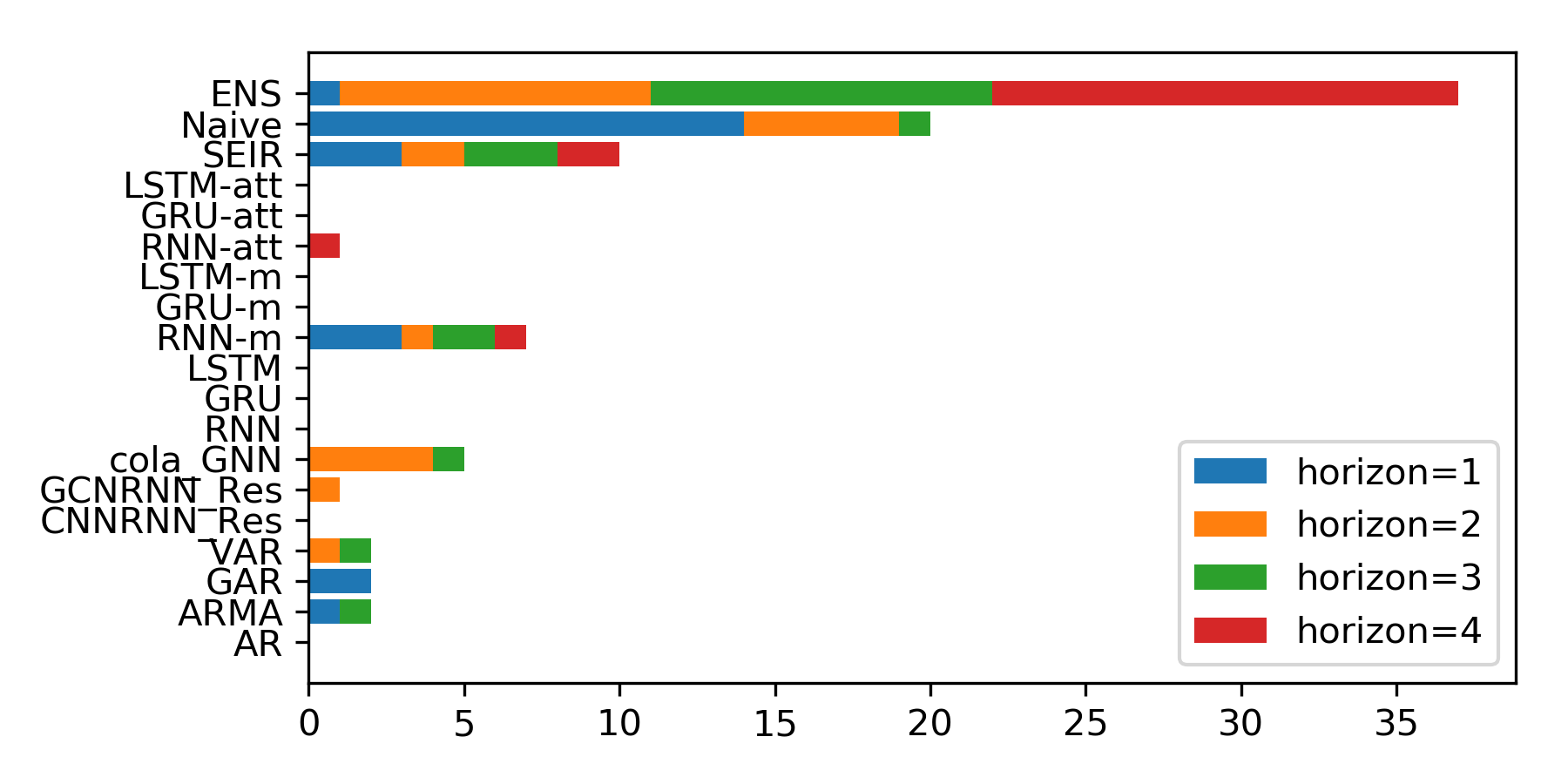}
    \subcaption{US-State}
    \label{fig:bar-us-state}
  \end{subfigure} 
  \quad
  \begin{subfigure}[b]{.3\textwidth}
    \centering
    \includegraphics[width=.93\textwidth]{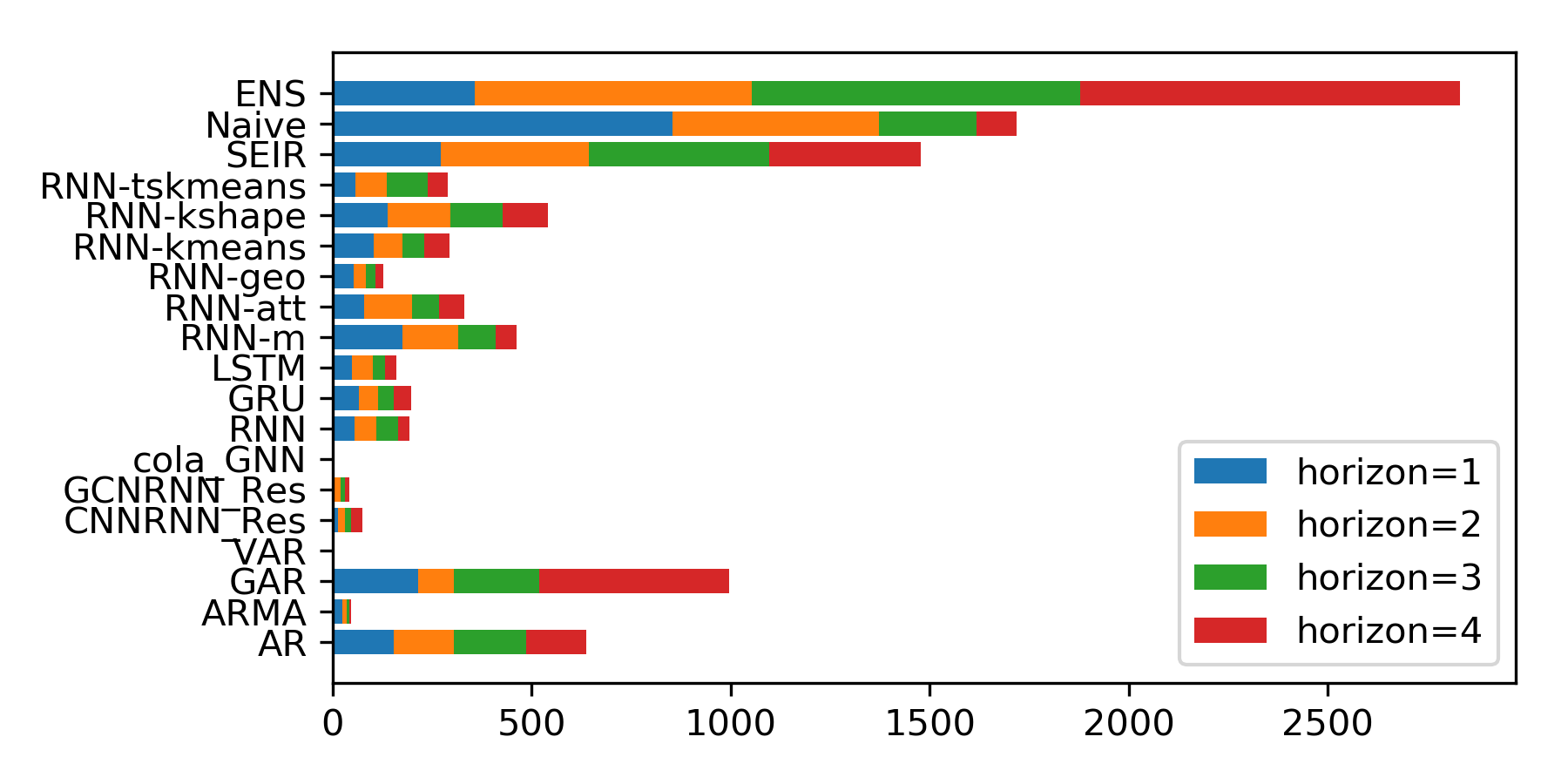}
    \subcaption{US-County}
    \label{fig:bar-us-county}
  \end{subfigure} 
  \caption{The distribution of best RMSE performance among individual methods. x-axis denotes the number of best performance achieved by each method.} 
  \label{fig:best-num}
  \vspace{-0.5cm}
 \end{figure*}

\subsection{Baselines}
To serve as baselines for comparing the individual models, we also implemented SEIR compartmental model and several statistical time series models as well as state-of-the-art deep learning models. There are a few deep learning models proposed recently for COVID-19 forecasting which have not been peer reviewed, thus we do not consider any models published within 2 months upon our completion of this paper. 
\begin{itemize}
\item \textbf{Naive} uses the observed value of the most recent week as the future prediction. 
\item \textbf{SEIR}~\cite{venkatramanan2017spatio} is an SEIR compartmental model for simulating epidemic spread. We calibrate model parameters based on surveillance data for each region. Predictions are made by persisting the current parameter values to the future time points and run simulations. 
\item \textbf{Autoregressive (AR)} uses observations from previous time steps as input to a regression equation to predict the value at the next time step. We train one model per region using AR order 3.
\item \textbf{Global Autoregression (GAR)} trains one global AR model using the data available from each region. This is similar to the clustering-based methods that we proposed in this paper. We train one model per resolution using AR order 3. 
\item \textbf{Vector Autoregression (VAR)} is a stochastic process model used to capture the linear interdependencies among multiple time series. We train one model per resolution using AR order 3.
\item \textbf{Autoregressive Moving Average (ARMA)}~\cite{contreras2003arima} is used to describe weakly stationary stochastic time series in terms of two polynomials for the autoregression (AR) and the moving average (MA). We set AR order to 3 and MA order to 2.  
\item \textbf{CNNRNN-Res}~\cite{wu2018deep} uses RNNs, CNNs, and residual links to capture spatio-temporal correlation within and between regions. We train one model per region. We set the residual window size as 3 and all the other parameters are set as the same as the original paper. 
\item \textbf{Cola-GNN} ~\cite{deng2019graph} uses attention-based graph neural networks to combine graph structures and time series features in a dynamic propagation process. We train one model per resolution. We set RNN window size as 3 and all the other parameters are set as the same as the original paper.
\end{itemize}

\subsection{Settings and Implementation Details}
We set training window size $T=3$ for all RNN-based models due to the short length of available CF and DT. We examine weekly CF forecasting at county and state level for US and country level for 8 countries of which at least one country is from each continent.  The forecasting is made to 1, 2, 3, 4 weeks ahead at each time point i.e. $h=\{1,2,3,4\}$. All RNN-based models consist of 2 recurrent neural network layers with 32 hidden units, 1 dense layer with 16 hidden units, 1 dropout layer with 0.2 drop probability. We set batch size as 32, epoch number as 500. Stacking ensemble model consists of 1 dense layer with 32 hidden units and RELU activation function. We train ensemble with batch size 8 and epoch number as 200. Adam optimizer with default settings and early stopping with patience of 50 epochs are used for all model training.  Geo-clustering and alg-clustering methods are applied when training county level models. We set the number of clusters for alg-clustering method as $k=50$. The clustering is conducted on the normalized training curves using MinMaxScaler. Single feature means time series of CF. For country level forecasting, $m$ features include CF, DT, CCGR, DCGR. For US state level forecasting, $m$ features include CF, DT, CCGR, DCGR, TR and TPR. And CF, DT, CCGR, and DCGR are included for US county level forecasting. AR-based models and CNNRNN-based models are trained with single feature time series. For all models, we run 50 Monte Carlo predictions. For SEIR method, we calibrate a weekly effective reproductive number ($\text{R}_{\text{eff}}$) using simulation optimization to match the new confirmed cases per 100k. We set the disease parameters as follows: mean incubation period 5.5 days, mean infectious period 5 days, delay from onset to confirmation 7 days and case ascertainment rate of 15\% \cite{lauer2020incubation}.

  \begin{figure}[t]
  \centering 
    \includegraphics[width=1.\linewidth]{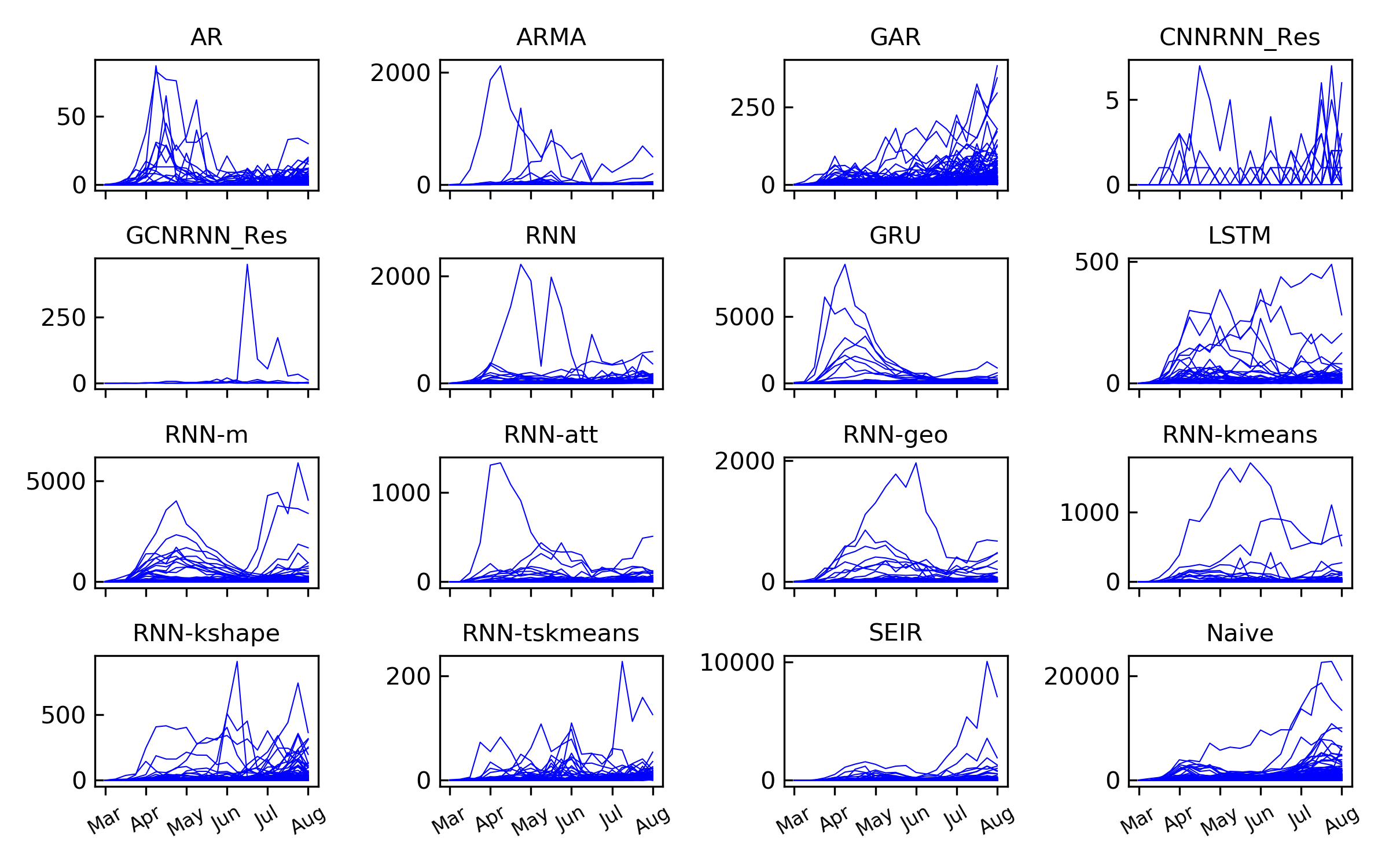}
  \caption{(US-county) The curves of weekly new confirmed cases grouped by individual models where the best RMSE performance is achieved. y-axis denotes new confirmed case count and x-axis denotes weeks (25 weeks). } 
  \label{fig:distribution}
  \vspace{-0.5cm}
 \end{figure}
 
\section{Results}
\subsection{Forecasting Performance}
We evaluate the model performance of horizon 1, 2, 3, and 4 at county-, state- and national-level using RMSE, MAPE and PCORR. To mitigate the performance bias caused by our settings, we divide the individual models into several categories based on different modules, training methods, features. Then we calculate the average performance per category. Note that an individual model may belong to multiple categories. 
\textbf{RNNs} includes models mainly consist of RNN module. 
\textbf{GRUs} includes  models mainly consist of GRU module.
\textbf{LSTMs} includes models mainly consist of LSTM module. 
\textbf{GNNRNNs} includes models mix CNN, RNN, GNN modules.
\textbf{ARs} includes autoregression based models.
\textbf{Vanillas} includes models in RNNs that use single feature and vanilla training.
\textbf{Clusters} includes models in RNNs that use single feature and geo, kmeans, tskmeans, kshape clustering training.
\textbf{SglFtrs} includes  \textit{RNN, GRU, LSTM}.
\textbf{MulFtrs} includes  \textit{RNN-m, GRU-m, LSTM-m, RNN-att, GRU-att, LSTM-att}.
\textbf{SEIRs} includes \textit{SEIR}.
\textbf{Naive} includes \textit{Naive}.
\textbf{ENS} is stacking ensemble of RNNs, GRUs and LSTMs.
GNNRNNs excludes \textit{cola-GNN} and ARs excludes \textit{VAR} for US-county forecasting due to their failures to make reasonable forecasting. For more details please refer to Table~\ref{table:res} note. 

Table~\ref{table:res} presents the numerical results. In general, we observe that (\textit{i}) at US state and county level ENS performs the best on 2, 3 and 4 weeks ahead forecasting while Naive performs the best on 1 week ahead. (\textit{ii}) SEIR outperforms others at global level forecasting on horizon 1, 2 and 3. (\textit{iii}) Models with a single type of DNN modules outperform those with mixed types of modules. (\textit{iv}) Models trained with vanilla methods outperform models trained with clustering-based methods. We will investigate and explain this observation in the next two paragraphs. (\textit{v}) Models trained with multiple features outperform models trained with a single feature at US state and county level. 

To better understand the model performance distribution over all regions, we select one individual method from each category without overlapping and count frequency of the best performance (\textbf{FRQBP}) per method. Fig.~\ref{fig:best-num} presents the aggregate counts of 1, 2, 3, 4 horizons. Note that methods with larger counts do not necessarily have better MAPE, RMSE and PCORR performance. The observations are in general consistent with those from Table~\ref{table:res} but with more specific observations regarding FRQBP: (\textit{vi}) the best 1 week ahead predictions are mostly achieved by Naive methods.  (\textit{vii}) For US state and county level, the best 2, 3, 4 weeks ahead predictions are achieved by ENS and the value increases as horizon increases.  (\textit{viii}) Alg-clustering-based models and models with multiple features achieve more best performance than vanilla models.  (\textit{ix}) GAR and AR have larger FRQBP than DNN models at US county level. 

Furthermore, in Fig.~\ref{fig:distribution} we show the US county level curves of weekly new confirmed cases grouped by individual methods where the best RMSE performance is achieved. It is interesting to observe that different methods achieve best performance over regions with different patterns, such as when the curves of weekly new confirmed cases have large fluctuation between subsequent weeks, the deep learning-based methods are able to capture the dynamics well as opposed to SEIR and Naive methods. The naive and SEIR models assume certain level of regularity in the time series, which tends to be violated in the curves pertaining to deep learning methods.
LSTM, RNN-kmeans, RNN-kshape, and RNN-tskmeans are outstanding in capturing dynamics with various patterns which show their generalization capability for time series forecasting. However, as we mentioned above the good performance in FRQBP does not indicate a better average performance on RMSE, MAPE, and PCORR since the latter also depends on the scales of ground truths. 
AR and GAR perform well on capturing dynamics of small number of cases.  
The CNNRNN-based methods does not perform well on county level forecasting. The likely reason is that the complexity of these models is much higher than simple RNN-based models and the complexity increases as the number of regions increases. Thus overfitting happens with such a small training data size at county level. 

We want to highlight that in order to investigate deep learning models for COVID-19 forecasting, the ensemble framework in this paper only combine DNN models. However it can but not necessarily include baselines like SEIR and Naive who perform very well in this task. We encourage researchers to ensemble models of various types to average the forecasting errors made by a particular poor model.

\begin{table*}[t]
\footnotesize
\centering
\caption{RMSE, MAPE and PCORR performance of different methods on the three datasets with horizon = 1, 2, 3, 4. Bold face indicates the best results of each column.}\label{table:res}
\scalebox{0.9}{\begin{tabular}{lcccccccccccc}
\toprule
 & \multicolumn{4}{c}{\textbf{Global}} & \multicolumn{4}{c}{\textbf{US-State}} & \multicolumn{4}{c}{\textbf{US-County}} \\
\cmidrule(r){2-5}\cmidrule(r){6-9}\cmidrule(r){10-13}
\textbf{ RMSE($\downarrow$) } & 1 & 2 & 3 & 4 & 1 & 2 & 3 & 4 & 1 & 2 & 3 & 4 \\
\midrule
ARs & 38067 & 46065 & 53942 & 57905         & 3255 & 3546 & 3822 & 4933         & 77 & 92 & 101 & 120 \\
CNNRNNs & 36895 & 49589 & 62499 & 69172         & 3511 & 4253 & 4615 & 5546         & 114 & 138 & 147 & 149 \\
RNNs & 31232 & 34877 & 44838 & 55403         & 2200 & 2940 & 3593 & 4605         & 60 & 80 & 96 & 110 \\
GRUs & 31172 & 36503 & 41513 & 55325         & 1936 & 2666 & 3520 & 4507         & 58 & 78 & 96 & 111 \\
LSTMs & 28023 & 35252 & 43130 & 53907         & 2031 & 2682 & 3576 & 4483         & 60 & 79 & 97 & 111 \\
Vanillas & 26323 & 33337 & 44273 & 54620         & 2135 & 2611 & 3415 & 4162         & 65 & 79 & 95 & 109 \\
Clusters & - & - & - & -         & - & - & - & -          & 72 & 91 & 103 & 117 \\
SglFtrs & 26878 & 33513 & 44838 & 54909         & 1824 & 2614 & 3533 & 4610         & 56 & 77 & 97 & 112 \\
MulFtrs & 32102 & 16588 & 42403 & 55019         & 1607 & 2231 & 3153 & 4110         & 50 & 68 & 85 & 99 \\
SEIRs & \textbf{8761} & \textbf{9393} & \textbf{13879} & 22805         & 2310 & 3362 & 4558 & 4635         & 65 & 75 & 82 & 96 \\
Naive & 15427 & 24899 & 27415 & 29318         & \textbf{1095} & 1936 & 1969 & 2466         & \textbf{37} & \textbf{48} & 60 & 71 \\
ENS & 18167 & 23203 & 28151 & \textbf{19559}         & 1261 & \textbf{1548} & \textbf{1598} & \textbf{2107}         & 46 & 49 & \textbf{59} & \textbf{62} \\
\midrule
\textbf{ MAPE($\downarrow$) } & 1 & 2 & 3 & 4 & 1 & 2 & 3 & 4 & 1 & 2 & 3 & 4 \\
\midrule
ARs & 173 & 167 & 187 & 195         & 2301 & 2571 & 1549 & 1821         & 129 & 119 & 121 & 127 \\
CNNRNNs & 95 & 123 & 173 & 197         & 1833 & 2656 & 1370 & 1777         & 148 & 187 & 202 & 191 \\
RNNs & 82 & 95 & 105 & 133         & 1265 & 1662 & 772 & 1084         & 116 & 142 & 153 & 162 \\
GRUs & 61 & 68 & 86 & 94         & 1335 & 1870 & 604 & 834         & 93 & 118 & 131 & 143 \\
LSTMs & 43 & 64 & 71 & 89         & 1453 & 1848 & 650 & 947         & 94 & 119 & 129 & 143 \\
Vanillas & 35 & 52 & 75 & 91         & 1092 & 1733 & 335 & 533         & 84 & 95 & 100 & 115 \\
Clusters & - & - & - & -         & - & - & - & -         & 140 & 167 & 171 & 179 \\
SglFtrs & 37 & 57 & 86 & 105         & 891 & 1260 & 509 & 719         & 94 & 122 & 139 & 152 \\
MulFtrs & 76 & 85 & 96 & 113         & 1450 & 1836 & 735 & 1103         & 103 & 130 & 138 & 143 \\
SEIRs & \textbf{12} & \textbf{12} & \textbf{18} & 28         & 996 & \textbf{1067} & 555 & 585         & 344 & 331 & 308 & 292 \\
Naive & 20 & 29 & 38 & 29         & \textbf{796} & 1198 & 565 & 590         & \textbf{75} & 98 & 95 & 83 \\
ENS & 26 & 29 & 31 & \textbf{23}         & 1049 & 1173 & \textbf{525} & \textbf{510}         & 90 & \textbf{95} & \textbf{91} & \textbf{81} \\
\midrule
\textbf{ PCORR($\uparrow$) } & 1 & 2 & 3 & 4 & 1 & 2 & 3 & 4 & 1 & 2 & 3 & 4 \\
\midrule
ARs & 0.8787 & 0.8335 & 0.8040 & 0.7995         & 0.8713 & 0.8257 & 0.7161 & 0.5214         & 0.7712 & 0.6070 & 0.5586 & 0.3062 \\
CNNRNNs & 0.9016 & 0.8479 & 0.8015 & 0.8217         & 0.7654 & 0.6441 & 0.5195 & 0.3119         & 0.1828 & -0.0232 & 0.0246 & 0.0636 \\
RNNs & 0.9477 & 0.9167 & 0.8690 & 0.7950         & 0.9094 & 0.8403 & 0.7974 & 0.6129         & 0.8321 & 0.7103 & 0.6086 & 0.5161 \\
GRUs & 0.9295 & 0.8968 & 0.8719 & 0.7966         & 0.9426 & 0.9152 & 0.8349 & 0.6791         & 0.8520 & 0.7377 & 0.5819 & 0.4776 \\
LSTMs & 0.9312 & 0.8829 & 0.8329 & 0.8030         & 0.9218 & 0.8776 & 0.7844 & 0.6782         & 0.8513 & 0.7226 & 0.5655 & 0.4779 \\
Vanillas & 0.9453 & 0.9106 & 0.8447 & 0.7703         & 0.9301 & 0.9094 & 0.8497 & 0.7521         & 0.8307 & 0.7528 & 0.6350 & 0.5297 \\
Clusters & - & - & - & -         & - & - & - & -         & 0.8167 & 0.6544 & 0.5242 & 0.4146 \\
SglFtrs & 0.9388 & 0.8989 & 0.8306 & 0.7560         & 0.9392 & 0.9035 & 0.8175 & 0.6635         & 0.8607 & 0.7347 & 0.5752 & 0.4744 \\
MulFtrs & 0.9351 & 0.8983 & 0.8707 & 0.8189         & \textbf{0.9660} & 0.9519 & 0.8960 & 0.7871         & 0.9293 & 0.8641 & 0.7669 & 0.7250 \\
SEIRs & \textbf{0.9957} & \textbf{0.9954} & \textbf{0.9851} & 0.9576         & 0.5806 & 0.5138 & 0.5379 & 0.3622         & 0.8632 & 0.7997 & 0.7809 & 0.7000 \\
Naive & 0.9888 & 0.9715 & 0.9498 & 0.9300         & 0.9764 & \textbf{0.9563} & 0.9208 & 0.8110         & \textbf{0.9546} & 0.9071 & 0.8485 & 0.7748 \\
ENS & 0.9661 & 0.9396 & 0.9162 & \textbf{0.9735}         & 0.9603 & 0.9487 & \textbf{0.9477} & \textbf{0.9070}         & 0.9159 & \textbf{0.9167} & \textbf{0.8620} & \textbf{0.8790} \\
\bottomrule
\end{tabular}}
\begin{tablenotes}
      \item \textbf{RNNs}: \textit{RNN, RNN-geo, RNN-m, RNN-att, RNN-kmeans, RNN-tskmeans, RNN-kshape}. 
      \textbf{GRUs}:  \textit{ GRU, GRU-m , GRU-att}.
      \textbf{LSTMs}:  \textit{ LSTM, LSTM-m, LSTM-att}.
      \textbf{GNNRNNs}:  \textit{ cola-GNN, GCNRNN-Res, CNNRNN-Res}.
      \textbf{ARs}:  \textit{ AR, ARMA, VAR, GAR}.
      \textbf{Vanillas}:  \textit{RNN}.
      \textbf{Clusters}:  \textit{RNN-geo, RNN-kmeans, RNN-tskmeans, RNN-kshape}.
      \textbf{SglFtrs}:  \textit{RNN, GRU, LSTM}.
      \textbf{MulFtrs}:  \textit{RNN-m, GRU-m, LSTM-m, RNN-att, GRU-att, LSTM-att}.
      \textbf{Naive}: \textit{naive}.
      \textbf{SEIRs}: \textit{SEIR}.
      \textbf{ENS} is stacking ensemble of the union of RNNs, GRUs, and LSTMs.
      CNNRNNs excludes \textit{cola-GNN} and ARs excludes \textit{VAR} for US-county forecasting due to their failures to make reasonable forecasting.  
\end{tablenotes}
\vspace{-0.5cm}
\end{table*}

\subsection{Sensitivity Analysis and Discussion}
In this section, we show sensitivity analysis on model types, feature number, and clustering method for individual models. 

\subsubsection{RNN modules}
\begin{figure*}[t]
  \centering 
   \begin{subfigure}[b]{.28\textwidth}
    \centering
    \includegraphics[width=1\textwidth]{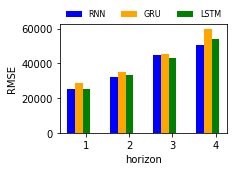}
    \subcaption{Global}
    \label{fig:model-global-rmse}
  \end{subfigure} 
  \quad
  \begin{subfigure}[b]{.28\textwidth}
    \centering
    \includegraphics[width=.97\textwidth]{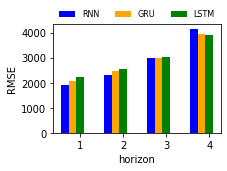}
    \subcaption{US-State}
    \label{fig:model-us-state-rmse}
  \end{subfigure} 
  \quad
  \begin{subfigure}[b]{.28\textwidth}
    \centering
    \includegraphics[width=.93\textwidth]{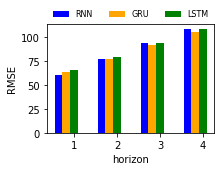}
    \subcaption{US-County}
    \label{fig:model-us-county-rmse}
  \end{subfigure} 
  \caption{Sensitivity analysis on RNN modules.} 
  \label{fig:sensitive-model}
  \vspace{-0.5cm}
 \end{figure*}

We compare RMSE performance of models with pure RNN, GRU, LSTM modules.
Fig.~\ref{fig:sensitive-model} shows the comparison between RNN, GRU, LSTM methods for three resolution datasets. We observe that RNN performs the best on 1 week ahead forecasting while GRU and LSTM outperform RNN on 3 and 4 weeks ahead forecasting at state and county level. The results indicate that RNN tends to perform better than GRU and LSTM for short-term forecasting while it loses advantage for long-term forecasting.  

\subsubsection{Number of features}
\begin{figure*}[t]
  \centering 
   \begin{subfigure}[b]{.28\textwidth}
    \centering
    \includegraphics[width=1\textwidth]{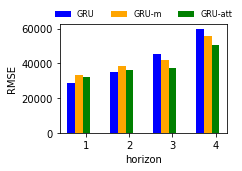}
    \subcaption{Global}
    \label{fig:feature-global-rmse}
  \end{subfigure} 
  \quad
  \begin{subfigure}[b]{.28\textwidth}
    \centering
    \includegraphics[width=.97\textwidth]{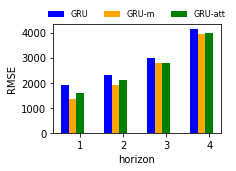}
    \subcaption{US-State}
    \label{fig:feature-us-state-rmse}
  \end{subfigure} 
  \quad
  \begin{subfigure}[b]{.28\textwidth}
    \centering
    \includegraphics[width=.95\textwidth]{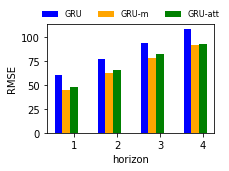}
    \subcaption{US-County}
    \label{fig:feature-us-county-rmse}
  \end{subfigure} 
  \caption{Sensitivity analysis on feature number.} 
  \label{fig:sensitive-feature}
  \vspace{-0.5cm}
 \end{figure*}

In our framework, we involve multiple data sources to model the co-evolution of multiple factors in epidemic spreading. We implement individual models either with single feature or with $m$ features. In addition, we use an attention layer to model the effect of other features on the target feature. Fig.~\ref{fig:sensitive-feature} presents the model performance of GRU, GRU-m, and GRU-att at three datasets. In general, GRU-m and GRU-att using $m$ features outperform GRU using single feature in most cases except for 1 and 2 week ahead forecasting at global level. Note that for global forecasting, there is no testing information which is a critical factor for revealing COVID-19 dynamics. 

\subsubsection{Clustering method}
\begin{figure*}[t]
  \centering 
   \begin{subfigure}[b]{.28\textwidth}
    \centering
    \includegraphics[width=1\textwidth]{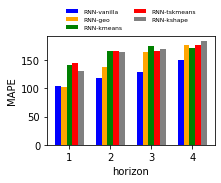}
    \subcaption{US-County MAPE}
    \label{fig:cluster-us-county-mape}
  \end{subfigure} 
  \quad
  \begin{subfigure}[b]{.28\textwidth}
    \centering
    \includegraphics[width=.97\textwidth]{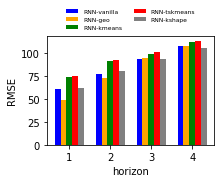}
    \subcaption{US-County RMSE}
    \label{fig:cluster-us-county-rmse}
  \end{subfigure} 
  \quad
  \begin{subfigure}[b]{.28\textwidth}
    \centering
    \includegraphics[width=.95\textwidth]{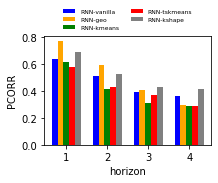}
    \subcaption{US-County PCORR}
    \label{fig:cluster-us-county-pcc}
  \end{subfigure} 
  \caption{Sensitivity analysis on clustering training.} 
  \label{fig:sensitive-cluster}
  \vspace{-0.5cm}
 \end{figure*}

Clustering-based training is applied in our framework to mitigate the likely overfitting due to small training data size. We compare US county level model performance of RNN, RNN-geo, RNN-kmeans, RNN-tskmeans, RNN-kshape. The comparison is shown in~\ref{fig:sensitive-cluster}. In general, we observe RNN, RNN-geo and RNN-kshape outperform RNN-kmeans and RNN-tskmeans. RNN-geo performs the best for 1 and 2 week ahead forecasting while RNN-kshape performs the best for 3 and 4 week ahead forecasting. This indicates that geo-clustering can capture near future co-evolution dynamics within a state informed by similar local epidemiological environments. Kshape clustering can further capture far future dynamics informed by other counties with similar trends. 

\section{Conclusion}
In this work, we developed an ensemble framework that combines multiple RNN-based deep learning models using multiple data sources for COVID-19 forecasting. The multiple data sources enable better forecasting performance. To mitigate the likely overfitting to noisy and small size of training datasets, we proposed clustering-based training method to further improve DNN model performance. We trained stacking ensembles to combine individual deep learning models of simple architectures. We show that the ensemble in general performs the best among baseline individual models for high resolution and long term forecasting like US state and county level. Ensembles play a very important role for improving model performance for COVID-19 forecasting. A comprehensive comparison between SEIR methods, DNN-based methods and AR-based methods are conducted. In the context of COVID-19, our experimental results show that different models are likely to perform best on different patterns of time series. 
Despite the lack of sufficient training data, DNN-based methods can capture the dynamics well and show strong generalization ability for high resolution forecasting as opposed to SEIR and Naive methods. Among multiple DNN-based models, spatio-temporal models are more likely to overfitting due to the high model complexity for high resolution forecasting. 

\section*{Acknowledgment}
The authors would like to thank members of the Biocomplexity COVID-19 Response Team and Network Systems Science and Advanced Computing (NSSAC) Division for their thoughtful comments and suggestions related to epidemic modeling and response support. We thank members of the Biocomplexity Institute and Initiative, University of Virginia for useful discussion and suggestions. 
This work was partially supported by National Institutes of Health (NIH) Grant R01GM109718, NSF BIG DATA Grant IIS-1633028, 
NSF Grant No.: OAC-1916805, NSF Expeditions in Computing Grant CCF-1918656, CCF-1917819, NSF RAPID CNS-2028004, NSF RAPID OAC-2027541,  US Centers for Disease Control and Prevention 75D30119C05935, DTRA subcontract/ARA S-D00189-15-TO-01-UVA. Any opinions, findings, and conclusions or recommendations expressed in this material are those of the author(s) and do not necessarily reflect the views of the funding agencies.

\bibliographystyle{IEEEtran}
\bibliography{ref.bib}

\end{document}